\documentclass[conference]{ieeeconf}
\usepackage{amsmath}
\usepackage{xcolor, soul}
\sethlcolor{yellow}
\usepackage{threeparttable}
\usepackage{multirow}
\usepackage{graphicx}
\usepackage{algorithm}
\usepackage{algpseudocode}
\usepackage{subcaption}
\usepackage{tabularx}
\usepackage{xcolor}
\usepackage{booktabs}
\usepackage{balance}
\usepackage[figurename=Fig.]{caption}

\usepackage{layouts}

\makeatletter
\let\NAT@parse\undefined
\makeatother
\usepackage{hyperref}
\hypersetup{
    pdfborder={0 0 0},
    colorlinks=true,
    urlcolor=teal,
    linkcolor=teal,
    citecolor=teal,
    bookmarks=true,
    bookmarksnumbered=true
}

\graphicspath{{../}}

\begin{document}

\title{PINN-Ray: A Physics-Informed Neural Network to Model Soft Robotic Fin Ray Fingers}

\author{Xing Wang$^{1}$, Joel Janek Dabrowski$^{2}$, Josh Pinskier$^{1}$, Lois Liow$^{1}$, Vinoth Viswanathan$^{1}$, \\Richard Scalzo$^{3}$, David Howard$^{1}$}

\maketitle

\begin{abstract}

Modelling complex deformation for soft robotics provides a guideline to understand their behaviour, leading to safe interaction with the environment. 
However, building a surrogate model with high accuracy and fast inference speed can be challenging for soft robotics due to the nonlinearity from complex geometry, large deformation, material nonlinearity etc. The reality gap from surrogate models also prevents their further deployment in the soft robotics domain.

In this study, we proposed a physics-informed Neural Networks (PINNs) named PINN-Ray to model complex deformation for a Fin Ray soft robotic gripper, which embeds the minimum potential energy principle from elastic mechanics and additional high-fidelity experimental data into the loss function of neural network for training. This method is significant in terms of its generalisation to complex geometry and robust to data scarcity as compared to other data-driven neural networks. Furthermore, it has been extensively evaluated to model the deformation of the Fin Ray finger under external actuation. 
PINN-Ray demonstrates improved accuracy as compared with Finite element modelling (FEM) after applying the data assimilation scheme to treat the sim-to-real gap.
Additionally, we introduced our automated framework to design, fabricate soft robotic fingers, and characterise their deformation by visual tracking, which provides a guideline for the fast prototype of soft robotics.

\end{abstract}


\IEEEpeerreviewmaketitle
\footnotetext[1]{ CSIRO Robotics, Data61, CSIRO, Australia}
\footnotetext[2]{ Statistical Machine Learning Group, Data61, CSIRO, Australia}
\footnotetext[3]{ Computational Modelling Group, Data61, CSIRO, Australia}

\section{Introduction}
Soft robotics concerns the design of robotic parts made from inherently compliant and low-stiffness materials. Owing to their flexibility, soft robotics offer safety advantages over rigid robotics when used to grasp or interact with fragile objects \cite{rus2015design}. 
However, the flexibility in soft robotics results in indefinite degrees of freedom (DoF) which makes it difficult to construct accurate and robust models that are able to achieve advanced tasks such as real-time control or computational design optimisation \cite{chen2020design}.

Extensive research has led to several approaches to improve accuracy and speed in soft robotics. These approaches can be classified into 3 main categories: physics-driven modelling \cite{qin2023modeling, armanini2023soft}, data-driven machine learning modelling \cite{yasa2023overview}, and hybrid physics-guided machine learning modelling \cite{mengaldo2022concise, sun2022physics}.

Physics-driven modelling models the soft robots with specific models, such as analytical geometry modelling \cite{fang2018geometry}, finite element modelling (FEM) \cite{wang2020soft}, and physical simulators \cite{collins2021review}.
Physics-rich analytical modelling using continuum mechanics or geometry modelling, can however be extremely challenging for soft robotics with complex geometry due to the nonlinear structure and material properties, and potentially nonlinear actuation mechanism (e.g. pneumatic air). This results in the sim-to-real gap where the model estimations can differ from reality and it is typically difficult to assimilate real-world data into the models to overcome this sim-to-real gap.

Data-driven modelling does not include any physics models but rather aims to fit machine learning models to large multi-fidelity datasets. Artificial neural networks are most commonly used and are applied to model soft robots' kinematics, static, and dynamics \cite{kim2021review}.  
Data-driven modelling is a mesh-free method but requires a large amount of training data to obtain a competent surrogate model. The experimental data can be expensive to obtain. Experimental data can be replaced or supplemented with simulation data, however this can be at the cost of accuracy due to the necessary approximation in the simulation parameters. Furthermore, although these models can close the sim-to-real gap, this is typically only applicable in proximity to where training data was provided. Without the physics, these models can tend to be inaccurate in scenarios where training data is scarce.

Recent developments combine physics-driven and data-driven modelling methods, leading to a physics-guided machine learning (PGML) \cite{karniadakis2021physics}. A highly promising approach is the physics-informed neural network (PINN), which embeds physical laws in the form of partial differential equations (PDEs) as constraints into the optimisation process of data-driven artificial neural networks. As the artificial neural network is trained to comply with the physical laws, it provides a robust surrogate model \cite{sun2021physics}. Furthermore, the PINN does not require large amounts of labelled data (if any) for training and this provides a significant advantage over purely data-driven modelling approaches.

        \begin{figure*}[!t]
            \centering
            \includegraphics[width=.95\linewidth]{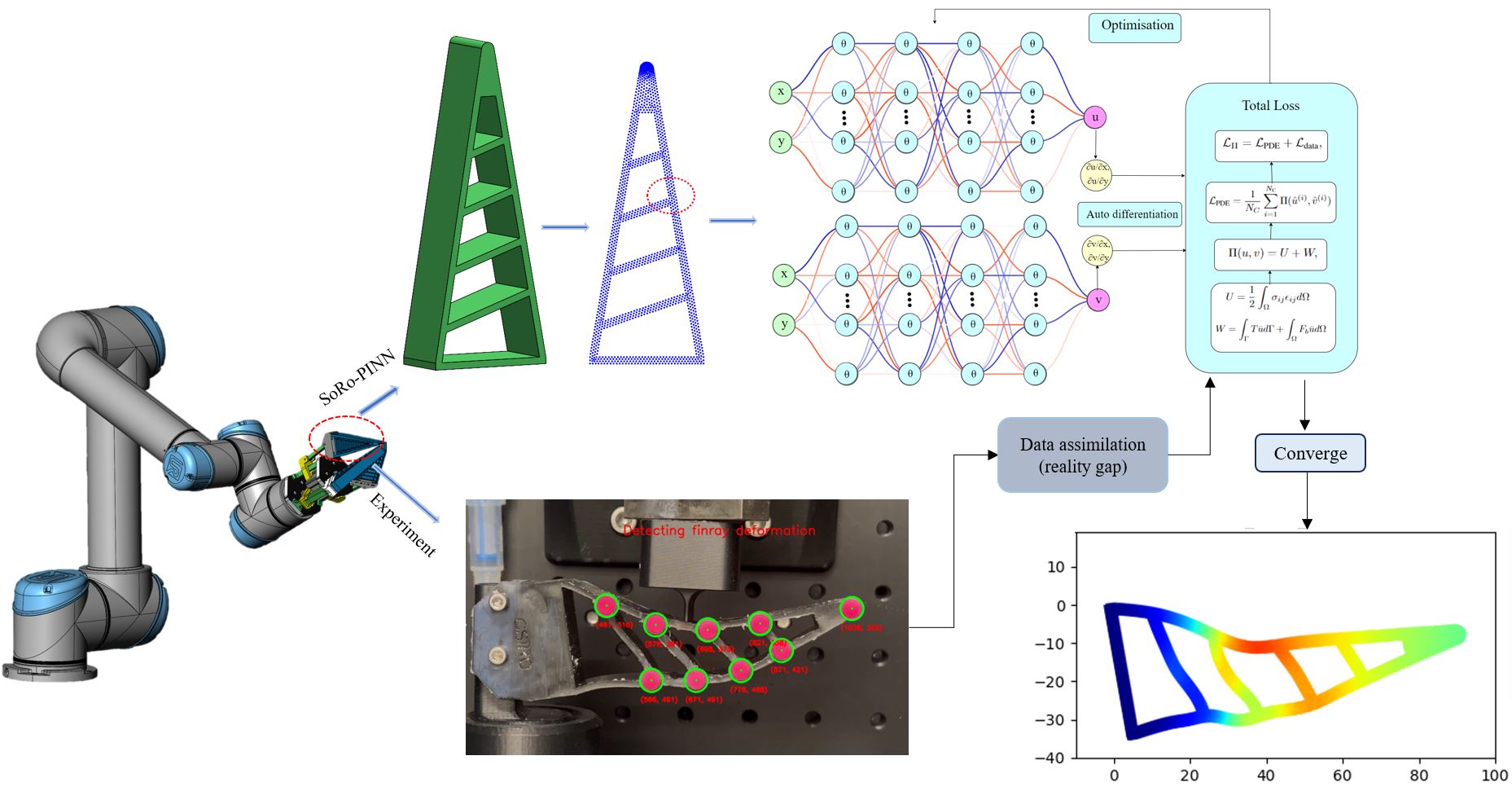}
            \caption{The framework of PINN-Ray to model the deformation of soft robotic finger with sim-to-real treatment. }
            \label{fig:pinn}
        \end{figure*}
In this work, we propose a machine learning-based modelling method to predict the deformation of soft robotics under external actuation, named PINN-Ray. The model is trained by incorporating the principle of minimum potential energy into the loss function, utilizing elastic mechanics theory. Additionally, we introduce data assimilation into the model where measurements from a real Fin Ray finger are incorporated into the training process. Our approach addresses the limitations of current learning-based methods by demonstrating robustness in scenarios with data scarcity. Two variations of PINN-Ray (one with and one without data assimilation) are employed to estimate the nonlinear displacement field of a Fin Ray finger and contrasted against the FEM. We show that our PINN-ray with data assimilation provides superior accuracy. Additionally, we present a real-time image processing technique that is employed to monitor the key points indicative of Fin Ray finger deformation for more efficient performance evaluation.
In summary, the contributions of this work are as follows:
    \begin{itemize}

        \item Present a PINN-Ray that combines physics from elastic mechanics theory and machine learning to predict the soft robotic deformation


        \item Provide a sim-to-real treatment by a modified loss function from experimental data assimilation 

        \item Demonstrate an image processing scheme for intrinsic key points tracking to efficiently characterise the soft robotic deformation
    
        \item Provide in-depth evaluation and comparison between PINN-Ray, FEM, and experimental results in modelling the soft robotic deformation 
        
    \end{itemize}
    The remainder of the paper is organized as follows: Section \ref{related} discusses the relevant modelling methods, then section \ref{method} introduces the detailed methodology. 
    The experiment results and discussion are presented in Section \ref{section: experiments}, followed by the conclusion in Section \ref{section: conclusion}. 

\section{Related work} \label{related}

   Extensive research efforts have been devoted to modeling the deformation, kinematics, statics, and dynamics of soft robotics.   
    Geometrical modeling stands out as a favored approach in physics-driven modeling for soft continuum robots, which offers a robust model to capture their shape variations, leveraging techniques such as constant curvature or specific geometric models. For example, Alici et al. modelled the deformation of a PneuNet soft robotic finger with the constant curvature assumption \cite{alici2018modeling}. Mbakop et al. utilised a Pythagorean Hodograph (PH) curves with prescribed lengths to reconstruct the robot shapes \cite{mbakop2021inverse}. 
    Nevertheless, the scope of geometric modeling is constrained by the physical shapes of soft robots, making it impractical for those with intricate geometries
    Continuum mechanics offers another avenue for modeling, driven by physics. It encompasses classic elastic and hyperelastic theories, which find extensive application in the field of soft robotics. For example, Trivedi et al. presented a model that accurately represents the large deformations and loading of a soft robotics manipulator, using the Neo-Hookean model for nonlinear elasticity and the Cosserat rod theory for the dynamics \cite{trivedi2008geometrically}. 
    The analytical approach using continuum mechanics requires precisely account for the specific geometry and shape of soft robots, making it challenging to generalize for other designs. This demands significant effort, even for seasoned researchers. 

     Data-driven modelling has enabled the creation of surrogate models in soft robotics, reducing the reliance on traditional physical laws.  Giorell et al. trained a data-driven neural network to model the cable-driven actuation force that can manipulate the soft robots to a certain end position. The training data were collected using cable tension, while an infrared camera was used to capture the tip positions. Extensive experimental data was acquired to for training and validation and residual between prediction and experimental data was minimised in the loss function  \cite{giorelli2015neural}. 
    However, data-driven modelling method is known for suffering from the challenges from data deficiencies. 
    
    The latest advancements in physics-informed machine learning have illuminated the modeling of soft robotics, representing a hybrid approach that integrates principles from both physics and data-driven machine learning.  
    Different strategies can be implemented to embed physical laws: train the data-driven machine learning with raw data generated by governing equations rather than collecting them from expensive and time-consuming experiments \cite{finegan2021application}; design a meticulous neural network architecture to formulate physical model \cite{zhang2018deepcg}, and lastly embed physical laws like continuum mechanics into loss function during neural network training \cite{bai2022introduction}. 
    There have been scant endeavors to apply PINN in the realm of soft robotics.
    Nava et al. proposed a novel and fully differentiable hybrid approach to fluid-structure interaction, which includes a U-net shaped physics-constrained neural network surrogate to simulate the flow field and then forward swimming of a soft-body fish \cite{nava2022fast}. PINN enables an accurate and efficient computational of swimming motion.
    Recently, Sun et al. added the physics laws including the elastic, viscous and Coulomb friction force to correct the pressure reading into recurrent neural network which was used to indirectly sensing the bending angle/force of the soft pneumatic actuators \cite{sun2022physics}. Their hybrid model significantly improve the prediction accuracy for soft robotics displacement. Finally, PINNs have been used to model wild fires with data assimilation and uncertainty quantification \cite{dabrowski2023bayesian}. The data assimilation improves the accuracy of the PINN by assimilating observations of fire-front locations into the model.
    

\section{Methodology}\label{method}

    \subsection{Elastic theory} 
    In this work, we make the following assumptions to the material property used for soft robotics: continuity, uniformity and isotropy. The constitutional material for soft robots are normally inherent compliant and flexible. 
    This study applied a linear elasticity material model to characterise the deformations.  
    
    Considering a Fin Ray finger design that occupies a two-dimensional spatial domain $\Omega$. Under external actuations, the Fin Ray finger will deform and its displacement field is given by $u$ in the horizontal or $x$-direction and $v$ in the vertical or $y$-direction \cite{shan2020modeling}. The total potential energy in the design domain is given by
    \begin{equation}
        \label{eq:totalEnergy}
        \Pi (u, v) = U + W,
    \end{equation}
    where $U$ is the total internal potential energy and $W$ is the potential energy from external actuation such as a force or moment.

    The internal energy is a result of stresses and strains in the body material and is given by
    \begin{equation}
        \label{eq:internalEnergy}
        U = \frac{1}{2} \int_{\Omega}  \sigma_{ij} \epsilon_{ij}  d \Omega,
    \end{equation}
    where $\sigma_{ij}$ is the stress tensor, $\varepsilon_{ij}$ is the strain tensor, and $i,j$ represent coordinate directions in index notation\footnote{Introductions to index notation can be found in standard elasticity theory books such as \cite{sadd2009elasticity}.}.
     
    The stress, strain, and deformation are related from the physical equations \cite{doghri2013mechanics}. Under static equilibrium conditions \cite{sadd2009elasticity}
    \begin{equation}
        \frac{\partial \sigma_{ij}}{\partial x_j} + f_i = 0,
    \end{equation}
    where $f_i$ is the body force. The strain tensor is given by
    \begin{equation}
        \label{eq:stress}
        \varepsilon_{ij} = \frac{1}{2} \left( \frac{\partial u_i}{\partial z_j} + \frac{\partial u_j}{\partial z_i} \right),
    \end{equation}
    where $\varepsilon_{ij}$ represents the strain tensor,  $u_i$ denotes the displacement component, and $z_j$ represents a dimension (e.g., $z_1 \equiv x$ and $z_2 \equiv y$ in the two-dimensional case). The stress tensor is given by
    \begin{equation}
        \label{eq:strain}
        \sigma_{ij} = C_{ijkl} \varepsilon_{kl}
    \end{equation}
    where $ C_{ijkl}$ is the stiffness tensor that depends on the body material and $i, j, k$, and $l$ are indices used for the index notation.

    In two dimensions, the constitution equations for plain strain are derived using Hooks law and can be simplified to the following \cite{timoshenko1970theory}
    \begin{align}
        \label{eq:hooksLawXX}
        \sigma_{xx} &=  \frac{E \mu}{(1+\mu) (1-\mu)}(\varepsilon_{xx} +  \varepsilon_{yy})+ \frac{E}{1+\mu} \varepsilon_{xx}, \\
        \label{eq:hooksLawYY}
        \sigma_{yy} &=  \frac{E \mu}{(1+\mu) (1-\mu)}(\varepsilon_{xx} +  \varepsilon_{yy})+ \frac{E}{1+\mu} \varepsilon_{yy}, \\
        \label{eq:hooksLawXY}
        \sigma_{xy} &= \frac{E}{1 + \mu} \varepsilon_{xy},
    \end{align}
    where $E$ is the Young’s modulus and $\mu$ is Poisson’s ratio. 

    Given the form of the stress equation in (\ref{eq:stress}) and strain equations in (\ref{eq:hooksLawXX}), (\ref{eq:hooksLawYY}) and (\ref{eq:hooksLawXY}), the total energy in equation (\ref{eq:totalEnergy}) is in the form of a partial differential equation (PDE) of the displacement field $(u, v)$ with respect to $x$ and $y$. The aim is to solve these equations to obtain the displacement field resulting from external actuation (under static equilibrium conditions).

    \subsection{physics-informed Neural Network}

     \begin{figure*}[!t]
        \centering
        \includegraphics[width=.8\textwidth]{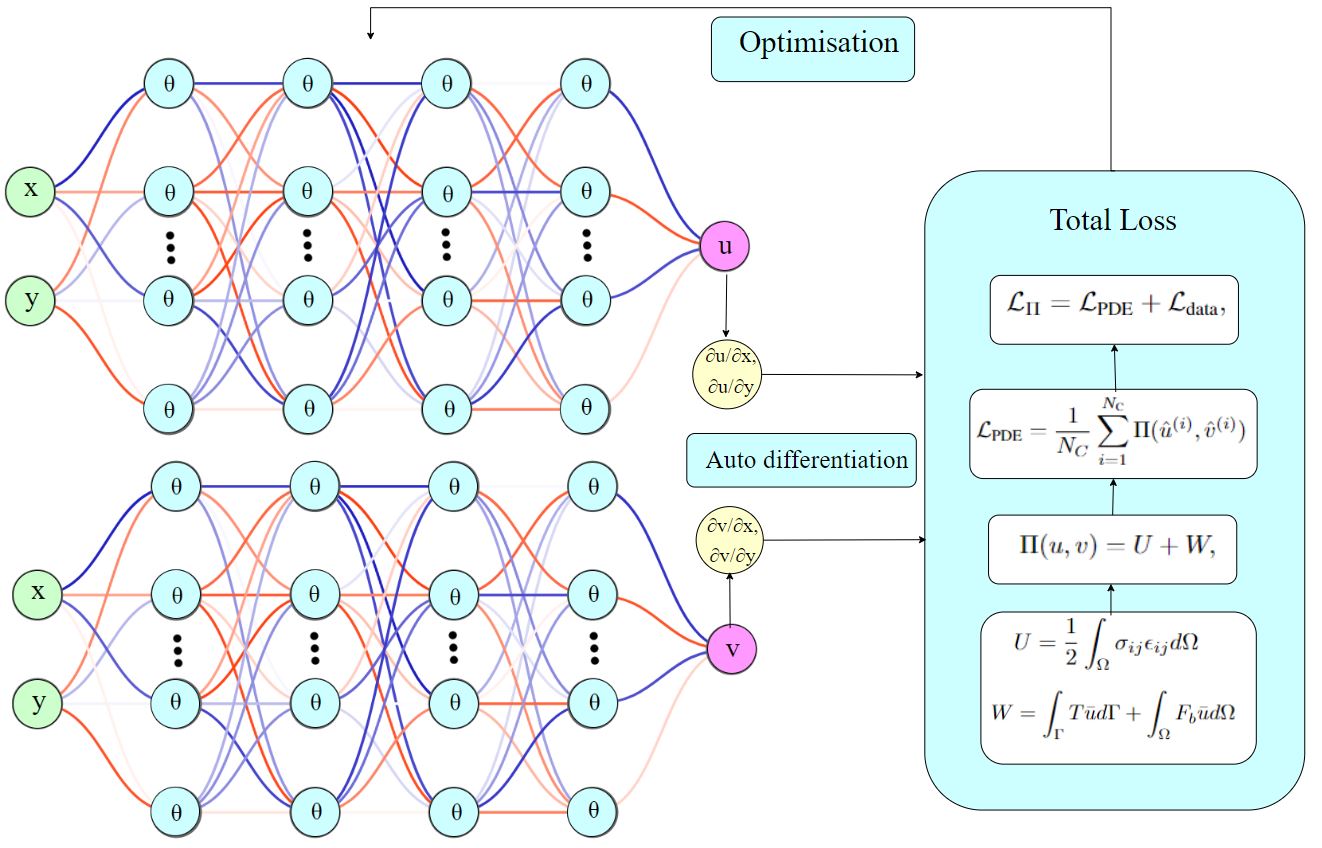}
        \caption{PINN-Ray architecture: two artificial neural networks are used to predict u and v respectively. The minimum potential energy method is implemented in the loss function}
        \label{fig:pinn}
    \end{figure*}

    The PINN approach to solving the PDE is formulate the problem as an optimisation problem. Under static equilibrium conditions, the total energy in the system given by (\ref{eq:totalEnergy}) should be zero. PINN-Ray uses a neural network $\psi(x, y; \theta)$ parameterised by $\theta$ to estimate the displacement field $(u, v)$ and optimise a neural network to minimise the total energy $\Phi(u,v)$ of the system. Let $\mathcal{L}_\Pi(\psi(x,y;\theta)$ be a loss function that represents the total energy in the system according to (\ref{eq:totalEnergy}) (see section \ref{sec:lossFunction}). The optimal parameters of the neural network are then given by
    \begin{align}
        \label{eq:optimisation}
        \theta^* = \arg \min_\theta \mathcal{L}_\Pi(\psi(x,y;\theta))
    \end{align}
    The neural network $\psi(x, y; \theta)$ takes the coordinates $(x, y)$ as inputs and predicts the displacement field $(u, v)$. Given that a neural network is differentiable, the partial derivatives of $u$ and $v$ with respect to $x$ and $y$ are computed using automatic differentiation (AutoDiff) \cite{baydin2018automatic}. Given these partial derivatives, the stress tensor in (\ref{eq:stress}) and the strain tensors in (\ref{eq:hooksLawXX}), (\ref{eq:hooksLawYY}) and (\ref{eq:hooksLawXY}) can be calculated and used to compute the total energy in (\ref{eq:totalEnergy}). Given the total energy, the neural network parameters can be optimised according to (\ref{eq:optimisation}). This process is illustrated in \figurename{} \ref{fig:pinn}.

    In this study, the fully connected neural network architecture proposed in \cite{wang2021understanding} is used. Given an input tensor $X$ (which is $(x,y)$ in this study), the output of an $L$-layered neural network can be computed as follows \cite{wang2021understanding}
    \begin{align*}
        & A = \phi \left( X W^1 + b^1 \right), \quad B = \phi \left( X W^2 + b^2 \right), \\
        & H^{(1)} = \phi \left( X W^{z,1} + b^{z,1} \right), \\
        & Z^{(k)} = \phi \left( H^{(k)} W^{z,k} + b^{z,k} \right), \quad k=1, \dots, L, \\
        & H^{(k+1)} = (1 - Z^{(k)}) \odot A + Z^{(k)} \odot B, \quad k=1, \dots, L, \\
        & \psi(X;\theta) = H^{(L+1)} W + b,
    \end{align*}
    where $\phi$ is an activation function, $W$ are weight matrices, and $b$ are bias vectors. The weight matrices and bias vectors are learnable parameters and are represented by $\theta$. 
    
    In this study, two neural networks $\psi_u(x,y; \theta_u)$ and $\psi_v(x,y; \theta_v)$ are used to separately output $u$ and $v$ respectively. One neural network could have been used to output a tensor comprising both $u$ and $v$, but we found that two separate neural networks performed better. We configure both neural networks to use the hyperbolic tangent function as an activation function. Furthermore, both neural networks comprise $L=4$ layers, where each layer contains 64 neurons.

    \subsection{Loss function}
    \label{sec:lossFunction}
    
    To perform optimisation in the neural network the energy is required to be represented in the form of a loss function. The PINN-Ray loss function will optimise over the total energy PDE, the boundary conditions, and data assimilation. This loss function is given by
    \begin{align}
        \mathcal{L}_\Pi = \mathcal{L}_\text{PDE} + \lambda_\text{BC} \mathcal{L}_\text{BC} + \lambda_\text{ASM} \mathcal{L}_\text{ASM},
    \end{align} 

    
    %
    where $\mathcal{L}_\text{PDE}$ is the total energy PDE loss, $\mathcal{L}_\text{BC}$ is the boundary condition loss, $\mathcal{L}_\text{ASM}$ is the data assimilation loss, and $\lambda_\text{BC}$ and $\lambda_\text{ASM}$ are hyper-parameters that specify the weighting of their respective loss component.
    
    Working backwards through the loss terms, for the data assimilation loss, we assume that a set of locations $(x_\text{ASM}^{(i)}, y_\text{ASM}^{(i)})$ with corresponding displacements $(u_\text{ASM}^{(i)}, v_\text{ASM}^{(i)})$, $i=1, \dots, N_\text{ASM}$ are available. These pairs are typically measured on a physical body. We can then construct the following Mean Squared Error (MSE) loss function
    \begin{align}
        \mathcal{L}_\text{ASM} = \frac{1}{N_\text{ASM}} \sum_{i=1}^{N_\text{ASM}}  
        & \bigg[
        \left( u_\text{ASM}^{(i)} - \psi_{u} (x_\text{ASM}^{(i)}, y_\text{ASM}^{(i)}) \right)^2 \nonumber \\
        & + \left( v_\text{ASM}^{(i)} - \psi_{v} (x_\text{ASM}^{(i)}, y_\text{ASM}^{(i)}) \right)^2  \bigg]
    \end{align}

    In the energy approach the free boundaries have natural boundary conditions, which are naturally satisfied \cite{sadd2009elasticity}. This leaves the essential boundary conditions such as fixed boundary conditions. Fixed boundaries are associated with no displacement and are thus Dirichlet boundary conditions where $u|_\text{fixed} = 0$ and $v|_\text{fixed} = 0$. In this study, we include an additional boundary condition associated with forcing, where a force along a boundary causes the boundary to displace by a specific known amount. Thus, given a set of locations $(x_\text{BC}^{(i)}, y_\text{BC}^{(i)}), ~ i = 1, \dots, N_\text{BC}$ along the fixed boundary and a set of locations along the forced boundary $(x_\text{F}^{(j)}, y_\text{F}^{(j)})$ with corresponding displacements $(u_\text{F}^{(j)}, v_\text{F}^{(j)})$, $j=1, \dots, N_\text{F}$. We thus construct the following MSE loss function
    \begin{align}
        \mathcal{L}_\text{BC} = 
        &\frac{1}{N_\text{BC}} \sum_{i=1}^{N_\text{BC}} \bigg[
        \left( \psi_{u} (x_\text{BC}^{(i)}, y_\text{BC}^{(i)}) \right)^2
        + \left( \psi_{v} (x_\text{BC}^{(i)}, y_\text{BC}^{(i)}) \right)^2 \bigg] \nonumber \\
        & + \frac{1}{N_\text{F}} \sum_{j=1}^{N_\text{F}} \bigg[ 
        \left( u_\text{F}^{(j)} - \psi_{u} (x_\text{F}^{(j)}, y_\text{F}^{(j)}) \right)^2 \nonumber \\
        & + \left( v_\text{F}^{(j)} - \psi_{v} (x_\text{F}^{(j)}, y_\text{F}^{(j)}) \right)^2 \bigg]
    \end{align}
    Finally, the total energy PDE loss is given by the MSE of (\ref{eq:totalEnergy}), which is computed using the neural network outputs and AutoDiff. In this particular study, we use a forced boundary condition in $\mathcal{L}_\text{BC}$ which accounts for the external energy applied to the system. We thus set $W=0$ in (\ref{eq:totalEnergy}) and the total energy is equivalent to the total internal energy $U$ in (\ref{eq:internalEnergy}). We approximate the integral in (\ref{eq:internalEnergy}) with a Monte Carlo estimate. For this, we sample a set of collocation points $(x_\text{C}^{(i)}, y_\text{C}^{(i)})$, $i= 1, \dots, N_\text{C}$ over the domain of $\Omega$, pass these through the neural network(s) to obtain $(\hat{u}^{(i)}, \hat{v}^{(i)})$ (which are the neural network estimates of $(u,v)$ at the given location). The PDE loss function is then given by
    \begin{align}
        \mathcal{L}_\text{PDE} = \frac{1}{N_C} \sum_{i=1}^{N_\text{C}} \Pi(\hat{u}^{(i)}, \hat{v}^{(i)})
    \end{align}

\section{Experiment }\label{section: experiments}

    \subsection{Design space \& manufacturing}
    
\begin{figure}[!b]
            \centering
            \includegraphics[width=.4\textwidth]{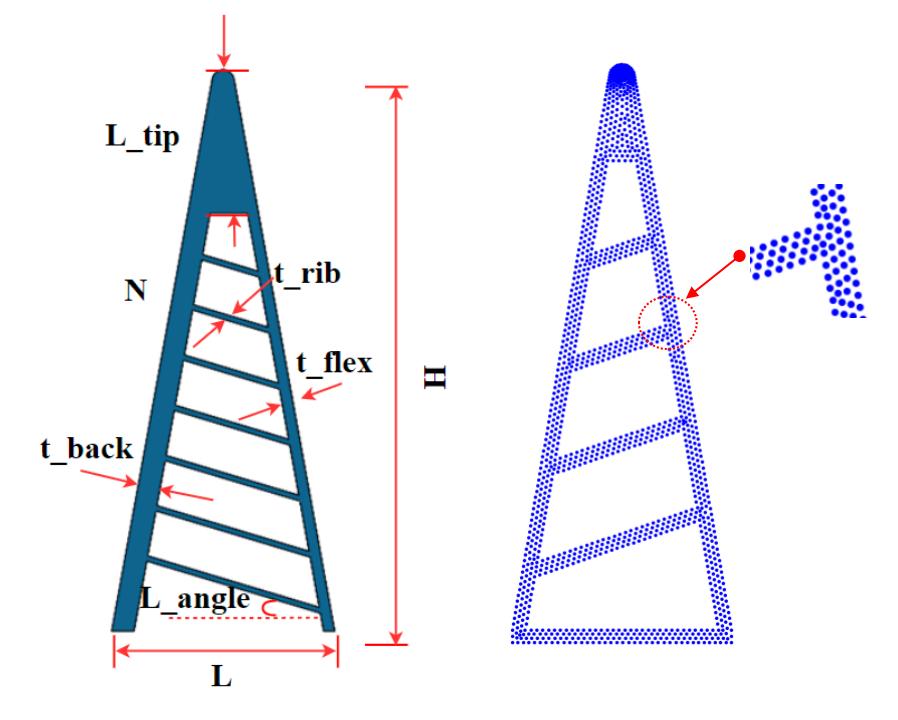}
            \vspace{-0.2cm}
            \caption{(a) Design space of the proposed soft robotics fingers, which is determined by 9 parameters, (b) Uniform sampling point discretization of the design space, while the design uses parameter values of [90 mm, 30 mm, 20 mm, 20 mm, 20$^0$, 4, 2 mm, 2 mm, 2 mm] }
            \label{fig:design}
        \end{figure}

    The robotic finger underwent design via geometrical construction using the Cadquery Python library, streamlined to be governed by 8 key parameters for 2D modeling  \cite{wang2024fin,xie2023fin}. Refer to the dimensions outlined in the accompanying \figurename{} \ref{fig:design} (a) for specifics. Then it was discretized into triangle elements using Gmsh library \cite{geuzaine2009gmsh}, where all the nodes were utilised as sampling points for training. 
    We numerically tested the Fin Ray finger design with design parameters $[H, L, W, L_{angle}, L_{tip}, N, t_{rib}, t_{flex}, t_{back} ]$, which have values of [90 mm, 30 mm, 20 mm, 20 $^\circ$, 20 mm, 4, 2 mm, 2 mm, 2mm], as illustrated in \figurename{} \ref{fig:design} (b).   The bottom edge is fixed as the boundary condition.
    We have assigned additional markers for further experimental deformation tracking, as shown in \figurename{ \ref{fig:jimstron}}.
    
    The robotic finger underwent one-shot 3D printing, where its CAD model was first sliced using GrabCAD. Next, the design was fabricated by the Stratasys J850 Polyjet printer, which seamlessly integrated markers into the process in a single step. Polyjet printer ejected liquid photopolymer into the building tray with multiple heads, which was then solidified under ultraviolet light. No support structure was enabled during the printing for a fast prototype.
    A digital material with shore hardness 85A was selected by digitally mixing two raw material: VeroUltra black and Agilus30 clear, whose material properties was determined using uniaxial tensile test \cite{pinskier2024diversity}. The properties of the material are such that Young's modulus is $E = 11.4$ MPa and Poisson's ratio is $\mu = 0.45$.

    \subsection{Experiment setup}

        \begin{figure}[!t]
            \centering
            \includegraphics[width=.45\textwidth]{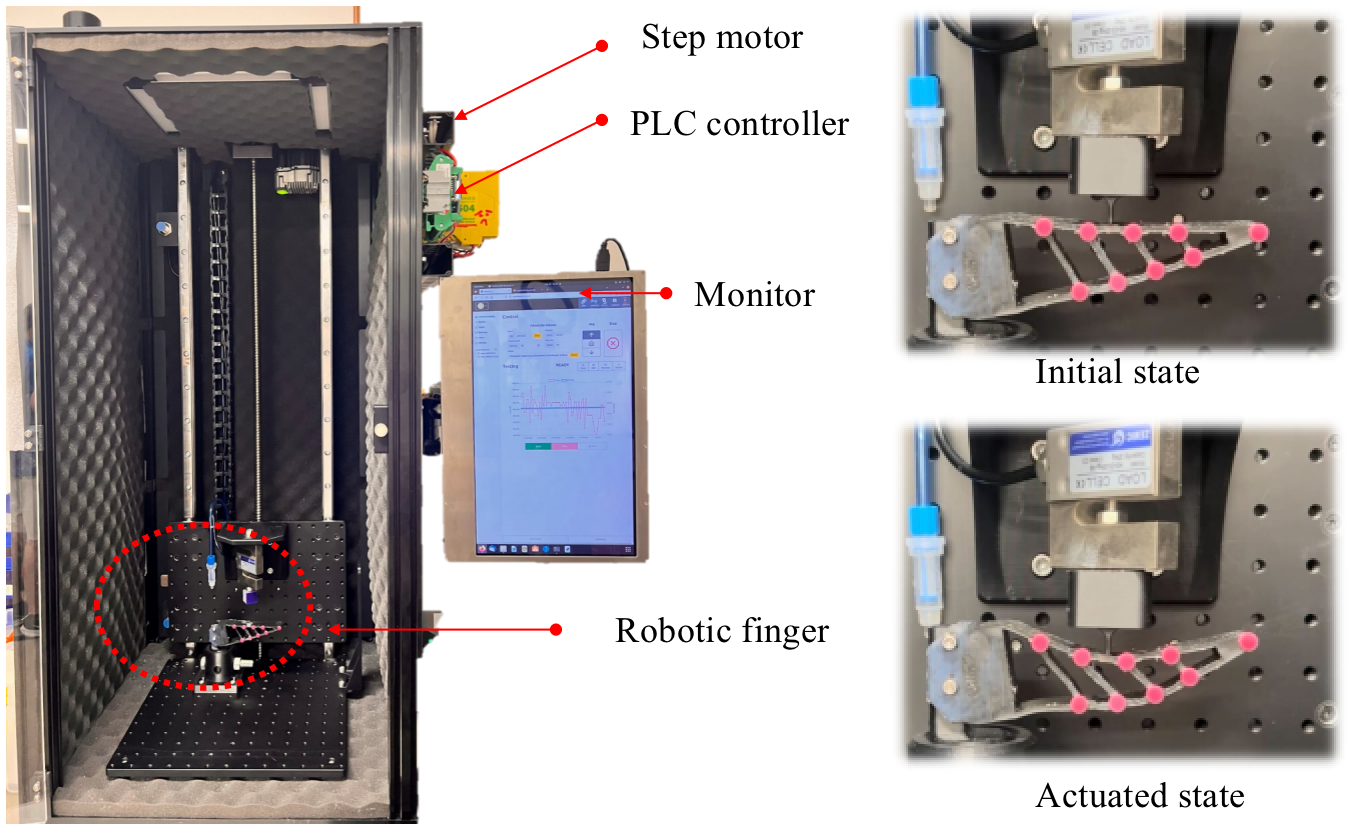}
            \caption{Left: Fin Ray finger under testing in the Jimstron test rig. Upper right: Fin Ray finger with no load. Lower right: Fin Ray finger under load. The markers are printed with robotic finger using a multi-material 3D printing step  }
            \label{fig:jimstron}
        \end{figure}
        
    To experimentally evaluate the deformation of the soft robotic finger, we used a customised automated testing rig named Jimstron, as shown in \figurename{} \ref{fig:jimstron}. The travel platform on Jimstron primarily undergoes vertical extension or contraction through the servo motor's action. At the end of this platform, we attached a 3D printed object for interacting with the soft robotic finger affixed to the base platform in the horizontal direction. 
    The rigid object moved toward the finger, causing a displacement upon contact, initiating deformation. 
    The fixed positions of the robotic finger were adjusted to enable contact points at various segments of the robotic finger.
    
    An RGB camera was positioned to capture the perspective view of the soft finger, which captured the deformation images of the soft finger after the travel platform reached to the preset distance. 
    The deformation was examined by monitoring the alterations in marker coordinates. To pinpoint their actual positions in the physical world, we set up a visual tracking system integrated with an image processing algorithm. We opted for an Intel web camera and pre-calibrated it using the Matlab Calibration Toolkit to acquire the intrinsic matrix.

        \begin{figure}[!t]
            \centering
            \includegraphics[width=.40\textwidth]{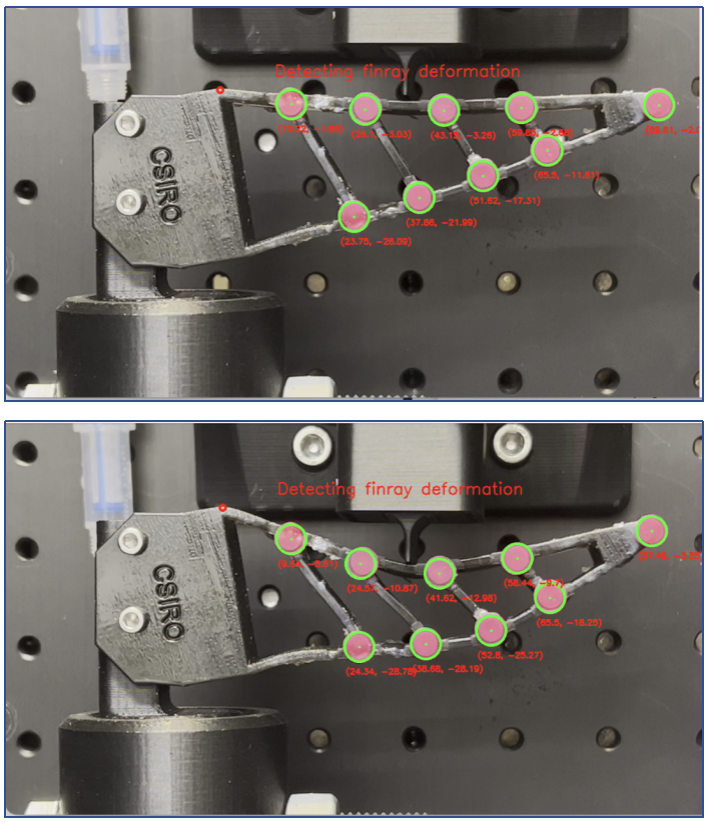}
            \caption{ Marker positions in pixel coordinates which are acquired by an image processing algorithm applied to images captured by a calibrated camera. Top: initial state; Bottom: actuated state}
            \label{fig:tracking}
            
        \end{figure}
        
    The movement of the finger was recorded within a 2D pixel frame and then translated into the designated global/world frame, which is attached to the corner of the Fin Ray finger as illustrated in \figurename{} \ref{fig:tracking}. The coordinate is in SI units as a fixed frame of reference. 
    An image processing algorithm was used to analyze the coordinates of all markers, extracting their color and shape characteristics, and locating their central positions within the pixel frame.
    The video stream frames were first converted to the BGR color space. Next, a medium blur with a size of 3 was applied to reduce noise. Following this, the frames were transformed into the CIELAB color space for color detection. Further geometry detection was achieved by applying a Gaussian blur operation, followed by Hough Circle detection.
    
    The detection positions were converted from pixel frame to world frame afterwards using a linear transform. The marker measurements at the initial and final static positions of the finger were were smoothed by averaging the measurements over several video frames while the finger remained in a static position. Given these smoothed measurements, the measured displacement is calculated by subtracting the marker positions at the final finger position from the marker positions at the initial finger position. The smoothed measured locations for the initial position denoted by $(x_\text{M}^{(i)}, y_\text{M}^{(i)})$ and the measured displacements are denoted by $(u_\text{M}^{(i)}, v_\text{M}^{(i)})$, where $i=1, \dots 9$. Note that the markers are indexed from left to right in \figurename{} \ref{fig:tracking}.
    
    As PINN-Ray is compared with the FEM, we use the FEM mesh to generate the collocation points $(x_\text{C}^{(i)}, y_\text{C}^{(i)})$, $i=1, \dots, N_\text{C}$ (randomly sampled locations over the domain $\Omega$ could also be used). A total of $N_\text{C} = 90959$ points are used. For the boundary along the base, a set of $N_\text{BC} = 1000$ evenly spaced locations are drawn from the base edge. For the forced boundary condition, the force is applied at location (35,0) to produce a displacement of (0mm, -10mm) and thus $N_\text{F}=1$ location point is drawn for this position. Finally, for the data assimilation, the measurements of the 9 markers at the fingers initial position are used and thus $N_\text{ASM}=9$.
        
    \section{Results \& Discussion}

%


We compare two variations of PINN-Ray to COMSOL FEM. The two variations are PINN-Ray with and without data assimilation and are referred to as PINN-Ray$_\text{asm}$ and PINN-Ray$_\text{std}$) respectively. Both PINN-Rays are trained over 60,000 epochs with a learning rate of $10^{-3}$, weight factors of $\lambda_\text{BC} = \lambda_\text{ASM} = 1000$, and using the ADAM optimisation approach. The learning rate and weight factors were determined by grid searches over the ranges $10^k$ where $k=-4, \dots, -2$ for the learning rate and $k=-1, \dots, 3$ for the weight factors. Larger weights for the boundary and assimilation loss functions ensured that these conditions were more strongly enforced.

To compare the methods, the Absolute Error (AE) and Mean Absolute Error (MAE) are used. For a given displacement estimate $(\hat{u}^{(i)}, \hat{v}^{(i)})$ at the location $(x_\text{M}^{(i)}, x_\text{M}^{(i)})$ and the corresponding measured displacements $(u_\text{M}^{(i)}, v_\text{M}^{(i)})$, the AE for marker $i$ is given by
\begin{align}
    \text{AE}^{(i)} = |\hat{u}^{(i)} - u_\text{M}^{(i)}| + |\hat{v}^{(i)} -  v_\text{M}^{(i)}|
\end{align}
The MAE is then given by
\begin{align}
    \text{MAE} =\frac{1}{N_{ASM}} \sum_{i=1}^9 \text{AE}^{(i)}
\end{align}

\begin{figure}[!t]
    \centering
    \begin{subfigure}{3.4in}
        \centering
        \includegraphics[width=\textwidth]{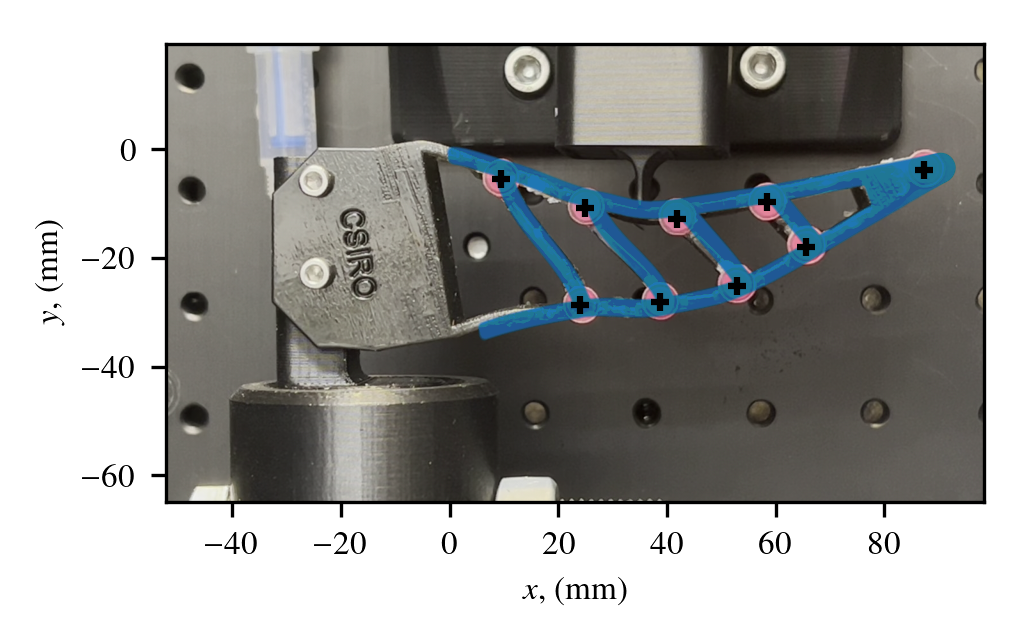}
        \vspace{-0.7cm}
        \caption{Finger displacement with an overlay of the FEM estimate.}
        \label{fig:assimFem}
    \end{subfigure}
    
    \begin{subfigure}{3.4in}
        \centering
        \includegraphics[width=\textwidth]{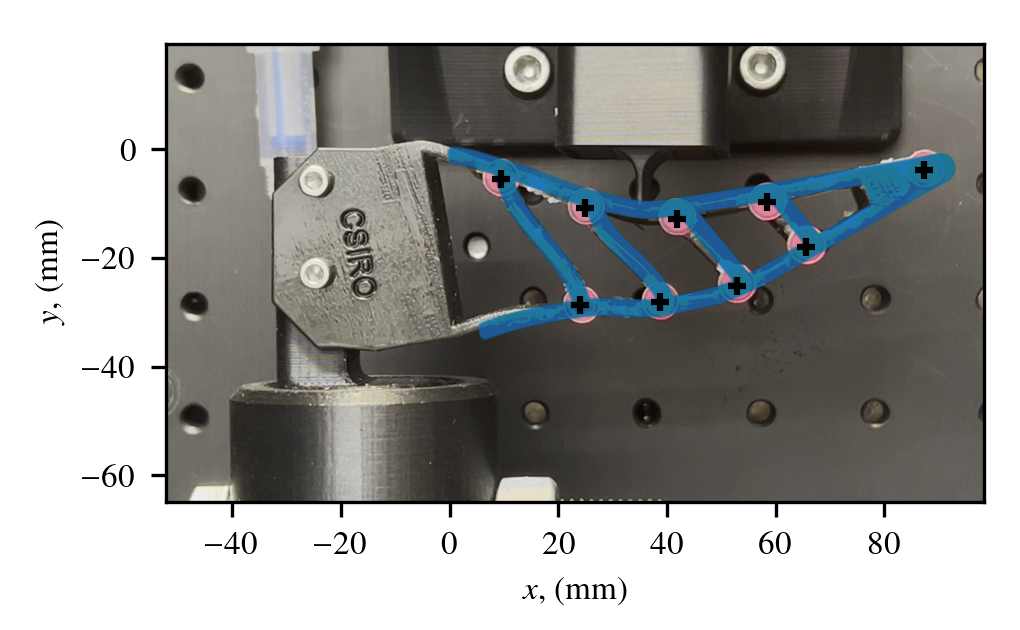}
        \vspace{-0.7cm}
        \caption{Finger displacement with an overlay of the PINN-Ray$_\text{std}$ estimate.}
        \label{fig:stdPinn}
    \end{subfigure}
    
    \begin{subfigure}{3.4in}
        \centering
        \includegraphics[width=\textwidth]{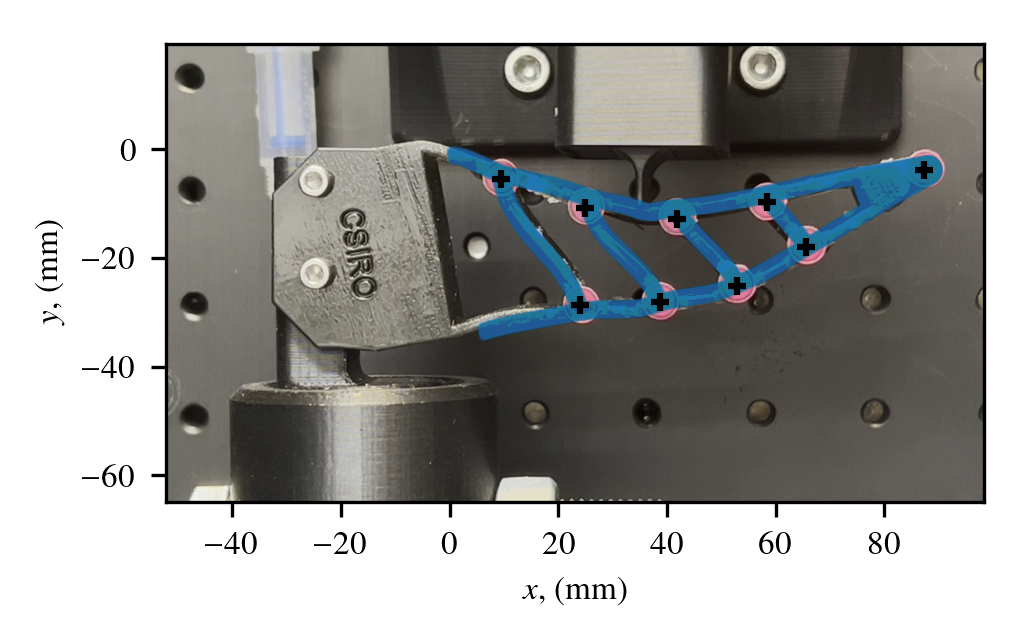}
        \vspace{-0.7cm}
        \caption{Finger displacement with an overlay of the PINN-Ray$_\text{asm}$ estimate.}
        \label{fig:assimPinn}
    \end{subfigure}
    
    \caption{Photographs of the displaced finger with the estimated displacements overlayed in blue. The black markers indicate the positions of the finger markers.}
    \label{fig:assimResults}
\end{figure}

\begin{figure}[!t]
    \centering
    \begin{subfigure}{3.4in}
        \centering
        \includegraphics[width=\textwidth]{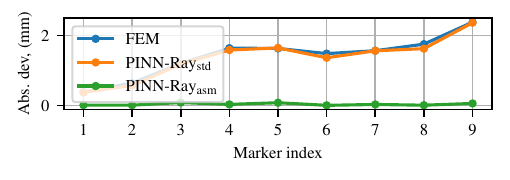}
        \vspace{-0.7cm}
        \caption{Absolute error for $u$.}
        \label{fig:assimAbsDevU}
    \end{subfigure}
    
    \begin{subfigure}{3.4in}
        \centering
        \includegraphics[width=\textwidth]{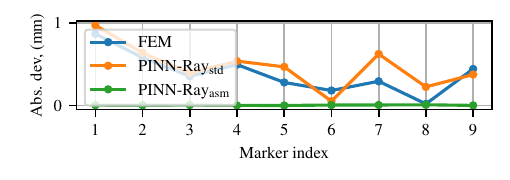}
        \vspace{-0.7cm}
        \caption{Absolute error for $v$.}
        \label{fig:assimAbsDevv}
    \end{subfigure}
    
    \caption{Comparisons of the absolute errors for each marker produced by the FEM and PINN-Ray$_\text{asm}$.}
    \label{fig:assimAbsDev}
\end{figure}

Illustrations of the actual displacements and the estimated displacements by the FEM, PINN-Ray$_\text{std}$, and PINN-Ray$_\text{asm}$ are provided in \figurename{} \ref{fig:assimResults}. The AE for these estimated displacements are illustrated in \figurename{} \ref{fig:assimAbsDev} and the MAE values are presented in Table \ref{table:mad}. The FEM and PINN-Ray$_\text{std}$ produce very similar results. PINN-Ray$_\text{std}$ is slightly less accurate for markers further away from the base (e.g., markers 7, 8 and 9). PINNs have been known to struggle to propagate information from boundary conditions further into the spatial domain and this could possibly be improved by increasing the number of training epochs. PINN-Ray$_\text{asm}$ achieves very low AE values for all markers and provides a significant improvement over both the FEM and PINN-Ray$_\text{std}$.

\begin{table}[!b]
    \centering
    \caption{MAE for the FEM, PINN-Ray$_\text{std}$, and PINN-Ray$_\text{asm}$}
    \label{table:mad}
    \begin{scriptsize}
        \begin{tabular}{ccccccccccc}
            \toprule
            & FEM & PINN-Ray$_\text{std}$ & PINN-Ray$_\text{asm}$ \\
            \midrule
            $u$ & 1.41 & 1.37 & 0.03 \\
            $v$ & 0.39 & 0.48 & 0.00 \\
            $disp$ & 1.463 & 1.452 & 0.03 \\
            \bottomrule
        \end{tabular}
    \end{scriptsize}
\end{table}

\begin{figure}[!t]
    \centering
    \includegraphics[width=3.4in]{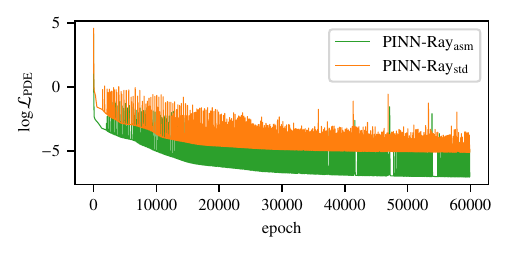}
    \vspace{-0.7cm}
    \caption{The log-loss ($\log \mathcal{L}_\text{PDE}$) over the training epochs.}
    \label{fig:loss}
\end{figure}

The log of the PDE loss ($\log \mathcal{L}_\text{PDE}$) for PINN-Ray$_\text{std}$ and PINN-Ray$_\text{asm}$ are plotted over the training epochs in \figurename{} \ref{fig:loss}. Including data assimilation significantly lowers the overall loss. With a lower loss, PINN-Ray$_\text{asm}$ has optimised to a lower energy level, which indicates that it has found a more accurate solution to the PDE. 

%

\begin{figure}[!t]
    \centering
    \includegraphics[width=3.4in]{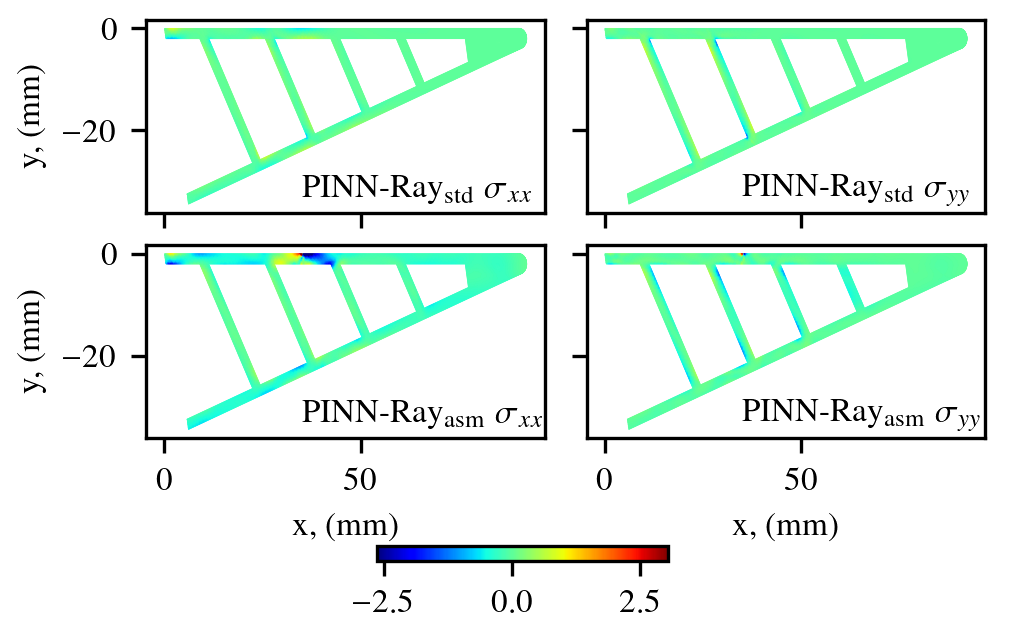}
    \caption{Stress estimates over the finger material.}
    \label{fig:stress}
\end{figure}

\begin{figure}[!t]
    \centering
    \includegraphics[width=3.4in]{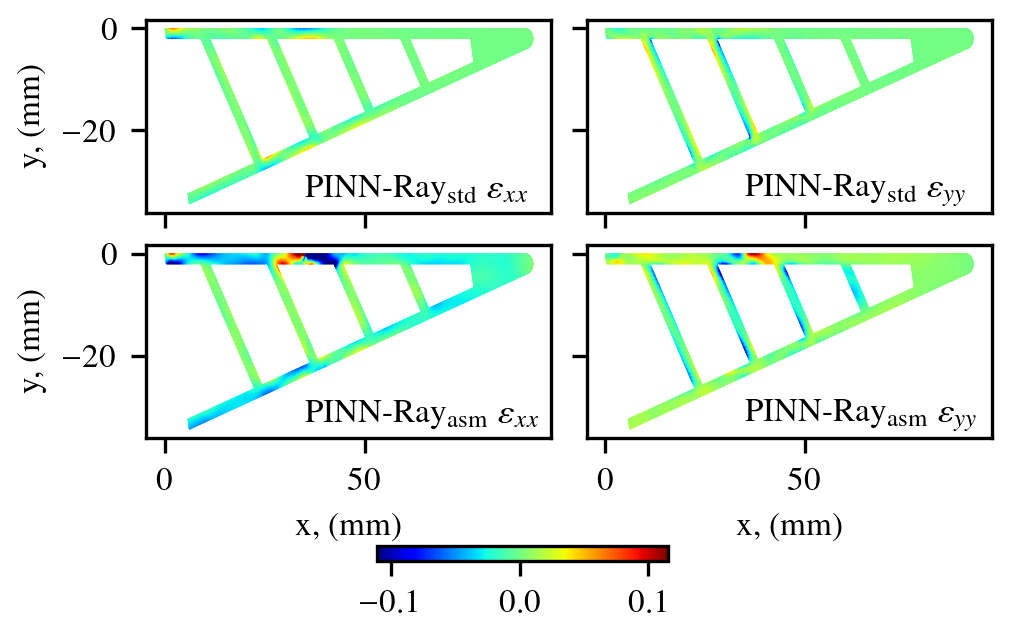}
    \caption{Strain estimates over the finger material.}
    \label{fig:strain}
\end{figure}

Finally, the stresses and strains in the material as estimated by PINN-Ray$_\text{std}$ and PINN-Ray$_\text{asm}$ are provided in Figs. \ref{fig:stress} and \ref{fig:strain}. As PINN-Ray$_\text{asm}$ has been shown to find a more optimal PDE solution, the estimates of the stresses and strains are likely to be more accurate. As shown in the figures, both the stresses and strains are underestimated by PINN-Ray$_\text{std}$. Underestimating the stresses and strains can be highly detrimental when a designer is considering yield points, elasticity, and toughness of the material. 

\section{Conclusion} \label{section: conclusion}
To address the challenges of present modelling methods for soft robotics, this work proposes a novel physics-informed neural network that combines physic laws and machine learning components, namely PINN-Ray. 
We demonstrate the detailed components of the network architecture, including the physical laws from elastic theory and nonlinear approximator artificial neural network. 
Most importantly, we provide a sim-to-real treatment involving experimental data assimilation to improve the modelling accuracy.
The proposed method is tested and validated on the classic Fin Ray robotic finger, whose deformation under external actuation is captured from FEM,  PINN-Ray$_\text{std}$, and PINN-Ray$_\text{asm}$. 

The MAE results suggest that PINN-Ray$_\text{std}$ achieves competitive accuracy compared with high-fidelity FEM.
PINN-Ray$_\text{asm}$ then significantly improves accuracy with experimental data assimilation ending up with an MAE of 0.03 as compared with 1.463 for the displacement, which validates its effectiveness as a treatment to bridge the sim-to-real gap for soft robotic deformation. Note that only 9 data points were assimilated to achieve these results.
Additionally, it improves the convergence speed and can provide more accurate estimations of latent variables such as stress and strain, which serve as the critical index to the durability evaluation of soft robotics design. 

\balance

\section*{Acknowledgement}
This work is supported by the Science and Industry Endowment Fund and CSIRO’s Future
Digital Manufacturing Fund.

\bibliographystyle{IEEEtran}
\bibliography{root}

\end{document}